\documentclass[11pt,onecolumn]{IEEEtran}

\usepackage[lmargin=1in, rmargin =1in, tmargin=1in,bmargin=0.5in]{geometry}
\usepackage{amsmath,amssymb,amscd}
\usepackage{graphicx}
\usepackage{cite}
\usepackage{subfigure}

\newtheorem{theorem}{Theorem}
\newtheorem{lemma}[theorem]{Lemma}

\newtheorem{corollary}[theorem]{Corollary}

\newtheorem{defn}[theorem]{Definition}

\newcommand{\R}{\mathbb{R}}
\newcommand{\ex}{\mathrm{e}}

\title{Curve Shortening and the Rendezvous Problem for Mobile Autonomous Robots}
\author{Stephen L. Smith, Mireille E. Broucke, and Bruce A. Francis%
\thanks{This work was in part supported by the National Sciences and Engineering Research Council of Canada (NSERC).}
\thanks{S. Smith is with the Department of Mechanical Engineering, University of California at Santa Barbara, Santa Barbara, CA 93106 USA (stephen@engineering.ucsb.edu).}
\thanks{M. Broucke and B. Francis are with the Department of Electrical and Computer Engineering, University of Toronto, ON, Canada, M5S 3G4 (broucke@control.utoronto.ca, bruce.francis@utoronto.ca).} }

\begin{document}
\maketitle

\begin{abstract}
If a smooth, closed, and embedded curve is deformed along its normal vector field at a rate proportional to its curvature, it shrinks to a circular point.  This curve evolution is called Euclidean curve shortening and the result is known as the Gage-Hamilton-Grayson Theorem.  Motivated by the rendezvous problem for mobile autonomous robots, we address the problem of creating a polygon shortening flow.  A linear scheme is proposed that exhibits several analogues to Euclidean curve shortening: The polygon shrinks to an elliptical point, convex polygons remain convex, and the perimeter of the polygon is monotonically decreasing.
\end{abstract}

\section{Introduction}
This paper studies the \emph{rendezvous problem} for mobile autonomous robots, in which the goal  is to develop a local control strategy that will drive each robots's state (usually its position) to a common value.  Research on this problem has been performed in discrete and continuous time.  The discrete time research can be split further into synchronous systems \cite{Ando_circum,bullo_rendezvous,ganguli_rendezvous,lin_morse,Moreau} (i.e., each robot moves only at global clock ticks), and asynchronous systems \cite{computer_science_1,computer_science_2} (i.e., no global clock is present).  In the synchronous case there have been several papers on \emph{circumcenter algorithms} \cite{Ando_circum,bullo_rendezvous,ganguli_rendezvous,lin_morse}, in which each robot moves towards the center of the smallest circle containing itself and every robot it sees.  In both the continuous and discrete time cases, the research has assumed fixed communication topologies---the sensors are omnidirectional and have a range larger than their environment, allowing each robot to see all others---and time-varying or state-dependent communication topologies---the sensors have limited range; the sensors are directional; or, communication links may be dropped or added.  In continuous time, a fair amount of research has been based on a simple strategy called \emph{cyclic pursuit} \cite{bruckstein,Josh,Zhiyun,Smith}.  In this strategy the agents are numbered from 1 to $n$, and each agent pursues the next with the $n$th agent pursuing the $1$st.

In this paper we look at the rendezvous problem from a different perspective.  We are concerned with the shape of the formation of robots as they converge to their meeting point.  We would like the formation to become more ``organized,'' in some sense, as time evolves. We use a simple model, numbering the robots from $1$ to $n$ and considering a fixed communication topology in continuous time.  We then view the robot's positions as the vertices of a polygon, and, motivated by the Gage-Hamilton-Grayson Theorem described below, we seek to create an analogous polygon shortening flow.

\begin{figure}
\begin{center}
\includegraphics[width=4.5cm]{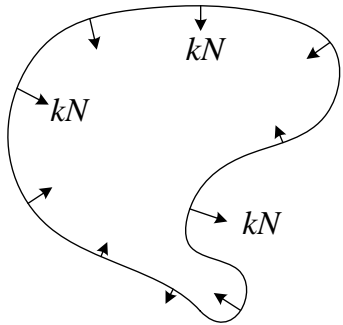}
\caption{The Euclidean curve shortening flow.}
\label{fig:curve_evolution}
\end{center}
\end{figure}

To introduce the Gage-Hamilton-Grayson Theorem, consider a smooth, closed curve $\mathbf{x}(p,t)$ evolving in time: $p\in[0, 1]$ parameterizes the curve; $t\geq 0$ is time; and $\mathbf{x}(p,t)\in \R^2$.  We can evolve this curve along its inner normal vector field $\mathbf{N}(p,t)$ at a rate proportional to its curvature  $k(p,t)$ (curvature is the inverse of the radius of the largest tangent circle to the curve at $\mathbf{x}(p,t)$, on the concave side):
\begin{equation}
\label{eq:euclidean_shortening}
\frac{\partial \mathbf{x}}{\partial t}(p,t) = k(p,t) \mathbf{N}(p,t).
\end{equation}
This curve evolution is known as the \emph{Euclidean curve shortening} flow \cite{curve_shortening}, and is depicted in Fig.~\ref{fig:curve_evolution}.
Let $L(t)$ and $A(t)$ denote respectively the length and enclosed area of the curve at time $t$.  Gage \cite{gage_1,gage_2,gage_hamilton}, Hamilton \cite{gage_hamilton}, and Grayson \cite{Grayson,As_fast_as_possible} showed that a smooth, closed and embedded curve evolving according to \eqref{eq:euclidean_shortening} remains embedded and shrinks to a circular point.  The term ``circular point" means that the curve collapses to a point and, if we zoom in on the curve as it is collapsing, the curve is becoming circular.  Throughout the evolution, $\dot A(t) = -2\pi$ and $L(t)$ is monotonically decreasing.  In \cite{As_fast_as_possible} it is also stated that under \eqref{eq:euclidean_shortening}, ``the curve is shrinking as fast as it can using only local information."  This notion will be clarified later.

There has been prior work in creating polygon shortening flows.  Motivated by the curve shortening theory, and applications in computer vision, Bruckstein et al. \cite{bruckstein} study the evolution of planar polygons in discrete time. A scheme is proposed that shrinks polygons to elliptical points (the vertices collapse to a point, and if we zoom in on the collapsing polygon, the vertices are converging to an ellipse).  In addition, \cite{bruckstein} discusses a polygon shortening scheme based on the Menger-Melnikov curvature \cite{Melnikov}.  In \cite{quadrilateral} this scheme is studied and it is shown that most quadrilaterals shrink to circular points.  In \cite{harvey} a flow is formulated such that the area enclosed by the polygon shrinks at a rate of $2\pi$ and the perimeter of the polygon is monotonically decreasing.

In this paper we study a planar polygon, with vertices $z_1,\ldots,z_n$ in the complex plane $\mathbb{C}$, as it evolves according to
\begin{equation}
\label{eq:evolution}
\dot z_i =\frac{1}{2}(z_{i+1}-z_i) + \frac{1}{2}(z_{i-1}-z_i), \quad i=1,\ldots,n,
\end{equation}
where the indices are evaluated modulo $n$.  Thus, vertex $i$ pursues the centroid (center of mass) of its two neighboring (according to numbering) vertices.  A discrete-time version of \eqref{eq:evolution} is studied in \cite{bruckstein}, and it is shown that the polygon shrinks to an elliptical point.  The contributions of this paper are as follows.  We introduce the curve shortening theory and its relation to the rendezvous problem.  We also demonstrate the importance of studying the shape of the formation of robots as they rendezvous.  We then show the following under \eqref{eq:evolution}:  1) convex polygons remain convex, 2) if vertices are arranged in a star formation about their centroid, they remain in a star formation for all time (in particular, the robots will not collide), 3) the perimeter of the polygon monotonically decreases to zero.  Finally, we derive the optimal direction for shortening the perimeter of a polygon.

\section{Polygon shortening}
\label{sec:polygon_shortening}
We consider $n$ robots in the plane to be the vertices of an $n$-sided polygon. In this section we formally define a polygon and introduce two polygon shortening schemes.

\subsection{Definition of an $n$-gon}
Following \cite{polygons} we introduce the definitions of a polygon and a simple polygon in $\R^2$ (or equivalently $\mathbb{C}$).  An \emph{$n$-gon} ($n$-sided polygon) is a (possibly intersecting) circuit of $n$ line segments $z_1z_2, z_2z_3, \ldots ,z_nz_1$, joining consecutive pairs of $n$ distinct points $z_1,z_2, \ldots, z_n$.  The segments are called \emph{sides} and the points are called \emph{vertices}.  A \emph{simple $n$-gon} is one that is non-self-intersecting. We denote the counterclockwise \emph{internal angle} between consecutive sides $z_iz_{i+1}$ and $z_{i-1}z_i$ of an $n$-gon as $\beta_i$ (as always, indices are modulo $n$).  For a simple $n$-gon these angles satisfy $\sum_{i=1}^n\beta_i=(n-2)\pi$.  An $n$-gon is \emph{convex} (\emph{strictly convex}) if it is simple and its internal angles all satisfy $0 < \beta_i \leq \pi$ ($0 < \beta_i < \pi$).

\subsection{Shortening by Menger-Melnikov curvature}
We now briefly describe the polygon shortening scheme studied in \cite{bruckstein,quadrilateral}, and our reasons for not following this approach.  Let $\mathbf{x}(p)$, $ p \in [0,1]$, be a smooth curve. Consider a set of parameter values $p_1 < p_2 < \cdots < p_n$ and the corresponding
discrete points $\mathbf{x}(p_i)$.  By connecting these points we create an $n$-gon.  As $n\rightarrow \infty$ and if the parameter values $\{ p_i \}$ become dense in $[0,1]$, the $n$-gon converges to the smooth curve $\mathbf{x}(p)$.  The idea is to create a polygon shortening scheme so that as $n\rightarrow \infty$, the scheme tends to \eqref{eq:euclidean_shortening}.

\begin{figure}
\begin{center}
\includegraphics[width=8cm]{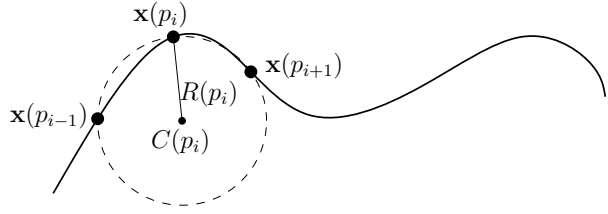}
\caption{The circumcenter for three points on the curve $\mathbf{x}(p)$.}
\label{fig:circumcenter}
\end{center}
\end{figure}

\begin{figure}
\begin{center}
\includegraphics[width=5cm]{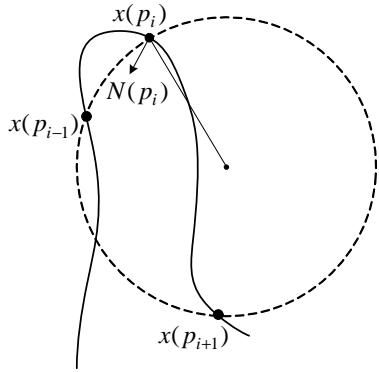}
\caption{The normal vector, and the Menger-Melnikov approximation, when the number of points $n$ is small.  The approximation to $\mathbf{N}(p_i)$ is very rough.}
\label{fig:convex_hull}
\end{center}
\end{figure}

If three consecutive points $\mathbf{x}(p_{i-1})$, $\mathbf{x}(p_{i})$, $\mathbf{x}(p_{i+1})$ are not collinear, there exists a unique circle (the \emph{circumcircle}) that passes through them.  Denote the radius of the circle by $R(p_i)$ and the center of this circle by $C(p_i)$, as shown in Fig. \ref{fig:circumcenter}.
The quantity $1/R(p_i)$ is called the \emph{Menger-Melnikov curvature} and has the property that
\[
\lim_{p_{i-1},p_{i+1}\rightarrow p_i}\frac{1}{R(p_i)}=|k(p_i)|.
\]
In addition, as the points $\mathbf{x}(p_{i-1})$ and $\mathbf{x}(p_{i+1})$ approach $\mathbf{x}(p_{i})$, the quantity $(C(p_i)-\mathbf{x}(p_i))/R(p_i)$ approaches $\mathbf{N}(p_i)$ if $k(p_i)>0$ and $-\mathbf{N}(p_i)$ if $k(p_i)<0$.  Therefore, we have
\[
\lim_{p_{i-1},p_{i+1}\rightarrow p_i} \frac{C(p_i)-\mathbf{x}(p_i)}{R(p_i)^2}=k(p_i)\mathbf{N}(p_i).
\]
The Menger-Melnikov flow is then given by
\[
\dot{\mathbf{x}}(p_i)=\frac{C(p_i)-\mathbf{x}(p_i)}{R(p_i)^2},  \ \ i=1,\ldots,n.
\]
This flow was studied in \cite{bruckstein,quadrilateral}.  However, due to the complexity of the system the results are quite limited \cite{bruckstein}.  In \cite{quadrilateral} it is shown that a simple $n$-gon collapses to a point in finite time, and for $n=4$ most quadrilaterals tend to regular polygons.  However when $n$ is small, this flow may yield a poor approximation of the inner normal vector, as shown in Fig.~\ref{fig:convex_hull}.  In fact, for a convex $n$-gon, the approximation to the normal vector may not even point into the interior of the $n$-gon.  Also, as the polygon collapses, the velocities of the vertices approach infinity, which is not ideal for our application.  In light of these remarks, we propose the scheme presented next.
\subsection{Linear scheme}
The linear polygon shortening scheme is given by \eqref{eq:evolution}. Defining the aggregate state $z=(z_1, \dots , z_n)$, we get the simple form $\dot{z}=Az$.  The matrix $A$ is \emph{circulant} (see Davis \cite{circulant}) and is given by $A=\mathrm{circ}\left(-1,\frac{1}{2},0,\ldots,0,\frac{1}{2} \right)$, with
\[
\mathrm{circ}(a_1,a_2,\ldots,a_n):=\begin{bmatrix}
a_1 & a_2 & \cdots & a_n \\
a_n & a_1 & \cdots & a_{n-1} \\
\vdots & \vdots & \vdots & \vdots \\
a_2 & a_3 & \cdots &  a_1 \\
\end{bmatrix}.
\]
The matrix $A$ can be written in terms of the polynomial
\[
q_A(s)=\frac{1}{2}s^{n-1}+\frac{1}{2}s-1,
\]
and the matrix $P=\mathrm{circ}(0,1,0,\ldots,0)$, as $A=q_A(P)$.  By the Spectral Mapping Theorem we obtain
\[
\mathrm{eigs}(A)=\{q_A(1),q_A(\omega),q_A(\omega^2),\ldots,q_A(\omega^{n-1}) \},
\]
where $\omega = \mathrm{e}^{2\pi j/n}$.  Therefore, denoting $\lambda_i:=q_A(\omega^{i-1})$, we have $\mathrm{eigs}(A)=\{\lambda_i:i=1,\ldots,n\}$.  Evaluating $\lambda_i$ we get
\begin{align*}
\lambda_i&=\frac{1}{2}\mathrm{e}^{2\pi j(n-1)(i-1)/n}+\frac{1}{2}\mathrm{e}^{2\pi j(i-1)/n}-1 \\
&=\cos(2\pi (i-1)/n)-1,
\end{align*}
where $i=1,\ldots,n$.  From this, one can easily verify the following properties: 1) the eigenvalues of $A$ are real, with one eigenvalue at zero, and all others on the negative real line, 2) the centroid  $\tilde z := \sum_{i=1}^nz_i/n$ is stationary throughout the evolution, and 3) the robots asymptotically converge to this stationary centroid.

The following theorem characterizes the geometrical shape of the points $z_i(t)$ as they converge to their centroid and is proved for discrete time in \cite{bruckstein}, and for general circulant pursuit in \cite{Josh_PhD}.
\begin{theorem}
\label{thm:ellipse}
Consider $n$ points, $z_1(t),\ldots,z_n(t)$ evolving according to \eqref{eq:evolution}.  As $t\rightarrow \infty$ these points converge to an ellipse.  That is, $z_1(t),\ldots,z_n(t)$ collapse to an  elliptical point.
\end{theorem}

\section{Star formations stay star formations}
\label{sec:star_formations}
\begin{figure}
\begin{center}
\includegraphics[width=8cm]{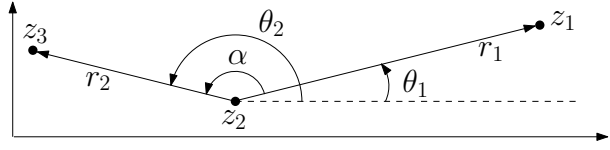}
\caption{The setup for the definition  of the function $F$.}
\label{fig:function}
\end{center}
\end{figure}

\begin{figure}
\begin{center}
\includegraphics[width=6cm]{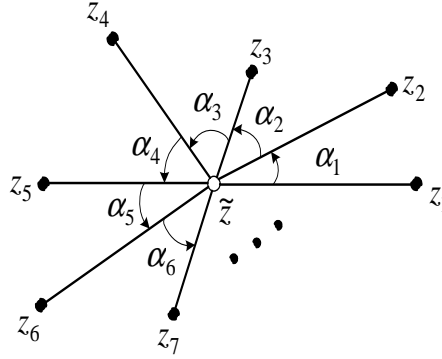}
\caption{A counterclockwise star formation.}
\label{fig:cc_star_formation}
\end{center}
\end{figure}

We will now see that if a group of robots have the initial arrangement shown in Fig. \ref{fig:cc_star_formation}, called a star formation, then under \eqref{eq:evolution}  it stays in a star formation and, in particular, no collisions occur.  We require some preliminary tools.  For $z\in\mathbb{C}$, we let $\Re\{z\}$, $\Im\{z\}$ and $\bar z$, denote the real part, imaginary part, and complex conjugate of $z$ respectively.
\begin{lemma}[Lin et al. \cite{Zhiyun}]
\label{lem:function}
Let $z_1$, $z_2$, and $z_3$ be three points in the complex plane, as shown in Fig. \ref{fig:function}.  Let $r_1:=|z_1-z_2|$, $r_2:=|z_3-z_2|$ and
\[
F=\Im \{ \overline{(z_1-z_2)}(z_3-z_2)\}.
\]
Then (i) $0 < \alpha < \pi$, $r_1 > 0$, and $r_2 > 0$ if and only if $F >0$;
(ii) $\pi < \alpha < 2\pi$, $r_1 > 0$, and $r_2 > 0$ if and only if $F <0$;
(iii) the points are collinear if and only if $F=0$.
\end{lemma}
\begin{proof}
Introduce the polar form
\[
z_1-z_2 =r_1\ex^{j\theta_1}, \quad z_3-z_2 =r_2\ex^{j\theta_2}
\]
where $\theta_1$, $\theta_2$ are the angles of the line segments in the global coordinate system.  Then
\[
F=\Im \{ \overline{(z_1-z_2)}(z_3-z_2)\} = \Im \{r_1\ex^{-j\theta_1}r_2\ex^{j\theta_2} \}= r_1r_2\sin(\alpha).
\]
Thus, $0 < \alpha < \pi$, $r_1 > 0$, and $r_2 > 0$ iff $F >0$; and $\pi < \alpha < 2\pi$, $r_1 > 0$, and $r_2 > 0$ iff $F <0$.  Also, the points are collinear iff $F=0$.
\end{proof}
Now consider our system of $n$ robots, whose positions, not all collinear, are denoted by $z_1,\ldots,z_n$.  Let $\tilde z$ be the centroid and $r_i$ be the distance from the centroid to $z_i$.  Let $\alpha_i$ denote the counterclockwise angle from $\tilde zz_i$ to $\tilde zz_{i+1}$ for $i=1,\ldots,n$, modulo $n$.

\begin{defn}[Lin et al. \cite{Zhiyun}]
The $n$ points are arranged in a \emph{counterclockwise star formation} if $r_i>0$ and $\alpha_i>0$, for all $i=1,\ldots,n$, and $\sum_{i=1}^n\alpha_i=2\pi$.  They are said to be arranged in a \emph{clockwise star formation} if $r_i>0$ and $\alpha_i<0$, for all $i=1,\ldots,n$, and $\sum_{i=1}^n\alpha_i=-2\pi$.
\end{defn}

This formation is shown in Fig. \ref{fig:cc_star_formation}.  In what follows we will consider only counterclockwise star formations, since the treatment for clockwise star formations is analogous.  Also, the case $n=2$ is trivial, so it is omitted.

We are now ready to state the main theorem of this section.
\begin{theorem}
\label{thm:star}
Suppose that $n$ distinct points, with $n >2$, are initially arranged in a counterclockwise star formation.  If these points evolve according to \eqref{eq:evolution} they will remain in a counterclockwise star formation for all time.
\end{theorem}

The proof uses these two results:

\begin{lemma}[Lin et al. \cite{Zhiyun}]
\label{lem:one_side_of_line}
Suppose that $n$ distinct points, $z_1,\ldots,z_n$, with $n > 2$ are in a counterclockwise star formation.  Then $\alpha_i < \pi$, $\forall i$.
\end{lemma}

\begin{lemma}[Lin et al. \cite{Zhiyun}]
\label{lem:collinear_4}
If $n$ points, $z_1,\ldots,z_n$ evolving according to $\eqref{eq:evolution}$ are collinear at some time $t_1$, then they are collinear for all $t <t_1$ and $t>t_1$.
\end{lemma}

{\em Proof of Theorem~\ref{thm:star}:} \
We begin by considering the function
\[
F_i(t) = \Im \{ \overline{(z_{i}(t)-\tilde z)}(z_{i+1}(t)-\tilde z)\}=r_{i}r_{i+1}\sin(\alpha_i).
\]
By the definition of a counterclockwise star formation we have $r_i(0) > 0$ and $0 < \alpha_i(0) <\pi$, $\forall i$.  Hence by Lemma \ref{lem:function}, $F_i(0) >0$, $\forall i$.  We want to show that $F_i(t) > 0$, $\forall i$ and $\forall t$, which by Lemma \ref{lem:function} shows that the vertices are in a counterclockwise star formation for all time.

Suppose by way of contradiction that $t_1$ is the first time that some $F_i$ becomes zero.  We can select $i=m$ such that $F_m(t_1)=0$ and $F_{m+1}(t_1) >0$, for if all the $F_i$'s are zero at $t_1$, then the points are collinear, which by Lemma \ref{lem:collinear_4} is a contradiction.  Hence, we have $F_i(t) > 0$ for all $t\in[0,t_1)$ and all $i$, $F_m(t_1) = 0$, and $F_{m+1}(t_1) > 0$.

Taking the time derivative of $F_m$, and noting that $\dot{\tilde z}=0$, we have $\dot F_m = \Im\{ \overline{\dot z_m}(z_{m+1}-\tilde z) + \overline{(z_{m} -\tilde z)}\dot z_{m+1}\}$.

By adding and subtracting $\tilde z$ in each term in \eqref{eq:evolution} we can write \eqref{eq:evolution} as
\[
\dot z_i=\frac{1}{2}(z_{i+1}-\tilde z) + \frac{1}{2}(z_{i-1}-\tilde z) +(\tilde z - z_i).
\]
Using this expression for $\dot z_m$ and $\dot z_{m+1}$ and simplifying, we obtain $\dot F_m = -2F_m +G_m$, where
\begin{align}
G_m&=\frac{1}{2}\Im\{\overline{(z_{m-1}-\tilde z)}(z_{m+1}-\tilde z)+ \overline{(z_{m}-\tilde z)}(z_{m+2}-\tilde z) \}\nonumber \\
\label{eq:G_m}
&=\frac{1}{2}\left(r_{m-1}r_{m+1}\sin(\alpha_{m-1}+\alpha_m) +r_m r_{m+2}\sin(\alpha_m+\alpha_{m+1}) \right).
\end{align}
Now, if $F_m(t_1)=0$, by Lemma \ref{lem:function}, one of the following four conditions must hold: (i) $\alpha_m(t_1)=\pi$ and $r_{m}(t_1),r_{m+1}(t_1) >0$; (ii) $\alpha_m(t_1)=0$ and $r_{m}(t_1),r_{m+1}(t_1) >0$; (iii) $r_m(t_1)= 0$;  (iv) $r_{m+1}(t_1)= 0$.

Condition (iv) cannot hold since $F_{m+1}(t_1)>0$.  Condition (i) cannot hold, for if it did, all points would lie on, or to one side of, the line formed by $z_{m+1}$ and $z_m$, a contradiction by either Lemma \ref{lem:one_side_of_line} or \ref{lem:collinear_4}.  Assume that condition (ii) holds.  Then $\alpha_m(t_1)=0$ and from \eqref{eq:G_m} we obtain
\begin{align*}
G_m(t_1) &= \frac{1}{2}\left(r_{m-1}r_{m+1}\sin(\alpha_{m-1}) +r_m r_{m+2}\sin(\alpha_{m+1}) \right) \\
 &=\frac{1}{2} \left(\frac{r_{m+1}}{r_{m}}F_{m-1}(t_1) +\frac{r_m}{r_{m+1}}F_{m+1}(t_1) \right).
\end{align*}
Since $r_m(t_1),r_{m+1}(t_1) >0$, $F_{m+1}(t_1) >0$, and $F_{m-1}(t_1) \geq 0$, it follows that $G_m(t_1) > 0$.  By continuity of $G_m$ there exists $0\leq t_0 < t_1$ such that $G_m(t) > 0$ for all $t\in[t_0,t_1]$.  Also, by assumption, $F_m(t) > 0$ for $t\in[0,t_1)$.  Therefore $\dot F_m(t) = -2F_m+G_m > -2F_m$ for all $t\in[t_0,t_1)$.  Integrating this and using the continuity of $F_m$, we obtain $F_m(t_1) \geq \ex^{-2(t_1-t_0)}F_m(t_0) > 0$,  a contradiction.

\begin{figure}
\begin{minipage}[b]{5cm}
\begin{center}
\includegraphics[width=5cm]{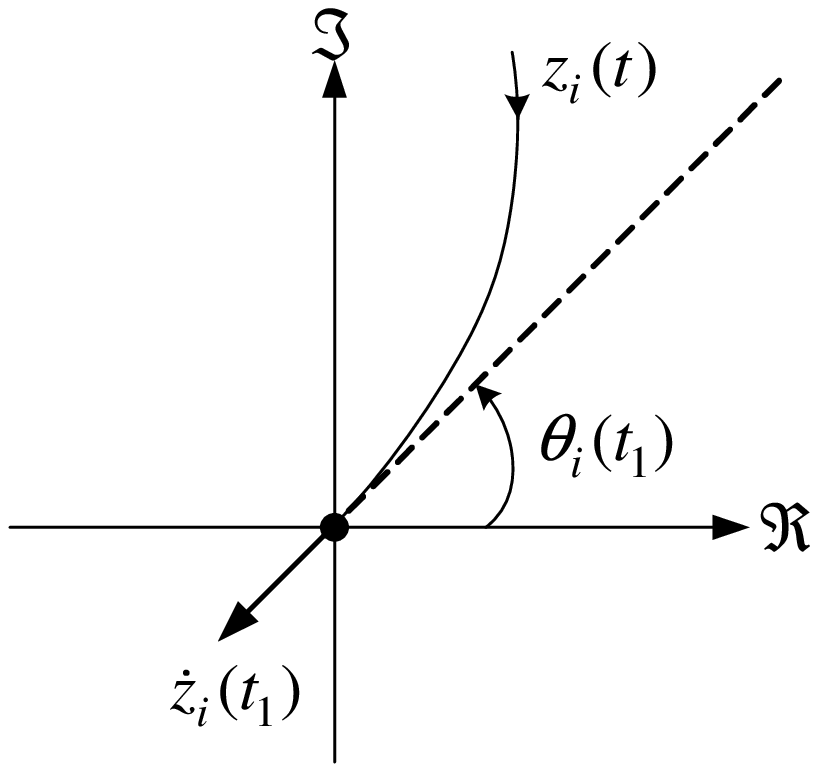}
\caption{The limiting $\theta_i(t)$ as $t\uparrow t_1$ when $z_i(t_1)=0$.}
\label{fig:limiting_theta}
\end{center}
\end{minipage}
\hfill
\begin{minipage}[b]{6cm}
\begin{center}
\includegraphics[width=5cm]{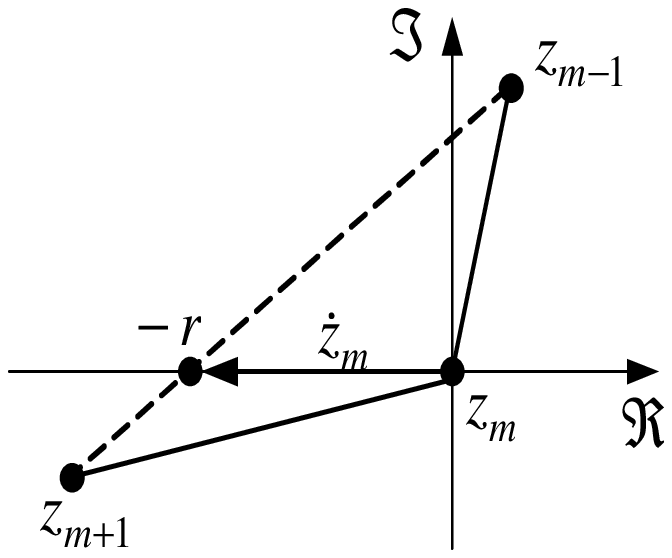}
\caption{The position of the points $z_{m-1}$, $z_{m}$, and $z_{m+1}$ at $t=t_1$.}
\label{fig:star_at_t1}
\end{center}
\end{minipage}
\end{figure}

Finally, suppose condition (iii) holds and $r_m(t_1)=0$.  Then $z_m(t_1)$ is positioned at the centroid, $\tilde z$.  Assume without loss of generality that $\tilde z=0$.  Notice that if $z_i(t_1)=0$, the angle $\theta_i(t_1)$ is not defined.
We now establish that if $z_i(t_1)=0$ and $\dot z_i(t_1) \neq 0$, then $\lim_{t\uparrow t_1}\theta_i(t)$ is well defined.  Expanding $z_i$ about $t_1$ we have $z_{i}(t_1)=z_i(t_1-h)+h\dot z_i(t_1) + {\cal O}(h^2)$, where ${\cal O}(h^2)/h \rightarrow 0$ as $h\rightarrow 0$.  If $z_i(t_1)=0$ then $z_i(t_1-h)=-h\dot z_i(t_1) + {\cal O}(h^2)$.  Hence, $\lim_{h\rightarrow 0} z_i(t_1-h)/h= -\dot z_i(t_1)$.  Therefore the limiting motion of $z_i(t)$ as $t\uparrow t_1$ is along the ray defined by $-\dot z_i(t_1)$, as shown in Fig.~\ref{fig:limiting_theta}.  Because of this, we can define
\begin{equation}
\label{eq:limit_theta}
\theta_i(t_1):= \left\{
\begin{aligned}
&\theta_i(t_1) \quad &\text{if} \quad r_i(t_1)> 0, \\
&\arctan\left( \frac{\Im\{-\dot z_i(t_1)\} }{\Re\{-\dot z_i(t_1)\} } \right) \quad &\text{if} \quad r_i(t_1)= 0.
\end{aligned}\right.
\end{equation}
With this definition we can talk about $\theta_i(t_1)$, and $\alpha_i(t_1)$, when $r_i(t_1)=0$.

Suppose that by a rotation of the coordinate system, if necessary, the vector $z_{m+1}(t_1)+z_{m-1}(t_1)$ lies on the negative real axis.  Then we can write
\begin{equation}
\label{eq:condition1}
\frac{z_{m+1}(t_1)+z_{m-1}(t_1)}{2} = -r, \quad \text{where}\quad r>0.
\end{equation}
We have $r>0$ for if $r=0$ then $z_{m-1}(t_1),z_m(t_1),z_{m+1}(t_1)$ all lie on a line through the centroid, and all other points must lie either on or to one side of this line, implying that 0 is not the centroid, or all the points are collinear, both contradictions.  Since $z_m(t_1)=0$, from \eqref{eq:evolution} and \eqref{eq:condition1} we have $\dot z_m(t_1)=-r$, as shown in Fig. \ref{fig:star_at_t1}.  If $n=3$ then $z_m(t_1)= 0$ and the centroid of $z_{m+1}(t_1)$ and $z_{m-1}(t_1)$ is at $-r$, implying that $0$ is not the centroid of the three points---a contradiction.

Therefore we need only consider $n>3$. Since $\dot z_m(t_1)=-r$, from \eqref{eq:limit_theta} we obtain
\begin{equation}
\label{eq:condition2}
\theta_m(t_1)=0.
\end{equation}
To obtain a contradiction for $n>3$ we will show that \eqref{eq:condition1} and \eqref{eq:condition2} cannot both be satisfied.  To do this we consider two cases, $r_{m-1}(t_1)=0$ and $r_{m-1}(t_1)>0$.  Since the points are in a star formation until $t_1$, we know that $\forall i$, $\alpha_{i}(t)\in(0,\pi)$ for $t\in[0,t_1)$.  Hence, if $\theta_i(t_1)$ and $\theta_{i+1}(t_1)$ are defined via \eqref{eq:limit_theta}, then by continuity, $\alpha_{i}(t_1)\in[0,\pi]$.

If $r_{m-1}(t_1)=0$ then from \eqref{eq:condition1} we have $z_{m+1}(t_1)=-2r$.  Therefore $\theta_{m+1}(t_1)=\pi$ and from \eqref{eq:condition2}, $\theta_m(t_1)=0$.  However this implies that all other $\theta_i(t_1)$'s that are defined must lie in $[-\pi,0]$.  Hence $\Im\{z_i(t_1)\}\leq 0$ $\forall i$, which implies that all points are collinear, or that 0 is not the centroid, both contradictions.

\begin{figure}
\begin{center}
\includegraphics[width=7cm]{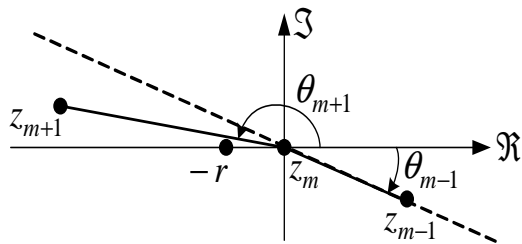}
\caption{The required geometry such that $\theta_{m-1}(t_1)\in[-\pi,0]$, $\theta_{m+1}(t_1)\in[0,\pi]$, and $z_{m+1}(t_1)+z_{m-1}(t_1) = -2r$.  All points lie either on or to one side of the dotted line.}
\label{fig:proof_region}
\end{center}
\end{figure}

If $r_{m-1}(t_1)>0$ then from \eqref{eq:condition2}, and since $\alpha_m(t_1),\alpha_{m-1}(t_1)\in[0,\pi]$, we have that $\theta_{m+1}(t_1)\in[0,\pi]$ and $\theta_{m-1}(t_1)\in[-\pi,0]$.  So $\Im\{z_{m+1}(t_1)\}\geq 0$ and $\Im\{z_{m-1}(t_1)\}\leq 0$. Because of this, as can be verified in Fig. \ref{fig:proof_region}, for \eqref{eq:condition1} to be satisfied either $z_{m-1}(t_1)$ and $z_{m+1}(t_1)$ are both real, in which case $\theta_{m+1}(t_1)-\theta_{m-1}(t_1)=\pi$, or neither is real and $\theta_{m+1}(t_1)-\theta_{m-1}(t_1)> \pi$.  But this implies that all points lie on, or to one side of, the line formed by $z_{m-1}(t_1)$.  Thus all points are collinear, or $0$ is not the centroid, both contradictions. \hfill $\Box$

\begin{figure}
\begin{center}
\includegraphics[width=10cm]{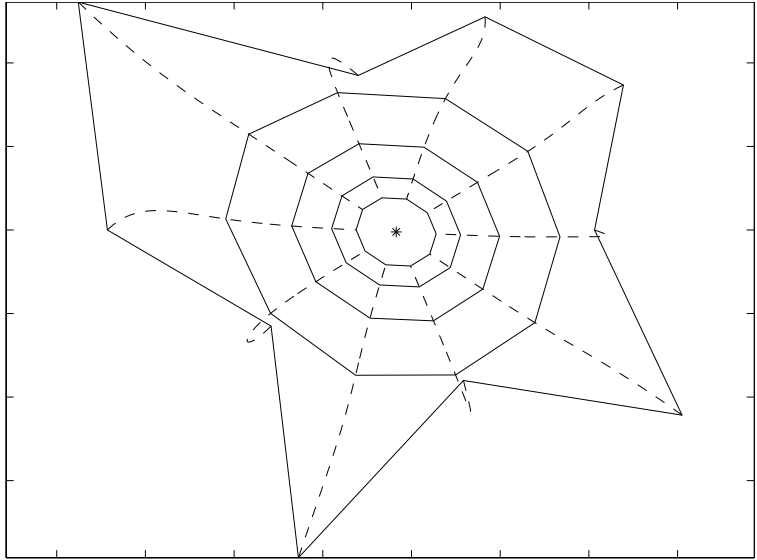}
\caption{The evolution of a polygon whose vertices are in a star formation about their centroid $*$.  The dashed lines show the trajectories of each vertex.}
\label{fig:star_evolution}
\end{center}
\end{figure}

\bigskip
Fig. \ref{fig:star_evolution} shows the evolution of a polygon that is in a star formation about its centroid.  Notice that the polygon remains in a star formation, becomes convex, and collapses to an elliptic point.

\section{Convex stays convex}
\label{sec:convex}
We now turn to the case where the formation is initially  a convex $n$-gon.  We will show that a convex $n$-gon evolving according to \eqref{eq:evolution} remains convex.  To do this we require a function similar to that in Lemma \ref{lem:function}, but which measures the counterclockwise internal angle between two sides of an $n$-gon.

\begin{figure}
\begin{center}
\includegraphics[width=8cm]{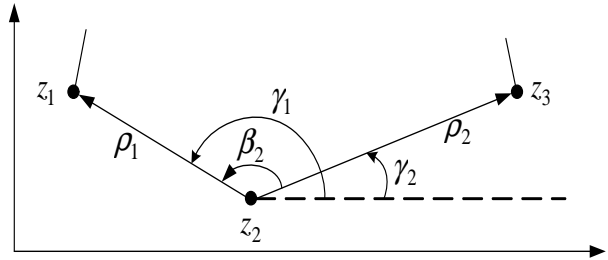}
\caption{The setup for the definition  of the function $H$.}
\label{fig:function2}
\end{center}
\end{figure}

\begin{lemma}
\label{lem:function2}
Consider a simple $n$-gon lying in the complex plane, whose vertices $z_i$ are numbered counterclockwise around the $n$-gon.  Let $\beta_2$ denote the counterclockwise angle from the side $z_2z_3$ to the side $z_1z_2$ as shown in Fig. \ref{fig:function2}, and define $\rho_1=|z_1-z_2|$, $\rho_2=|z_3-z_2|$ and
\[
H=\Im \{ (z_1-z_2)\overline{(z_3-z_2)}\}.
\]
Then (i) $0 < \beta_2 < \pi$, $\rho_1 > 0$, and $\rho_2 > 0$ if and only if $H >0$.
(ii) $\pi < \beta_2 < 2\pi$, $\rho_1 > 0$, and $\rho_2 > 0$ if and only if $H <0$.
(iii) the points are collinear if and only if $H=0$.
\end{lemma}
\begin{proof}
We introduce the polar form:
\[
z_1-z_2 =\rho_1\ex^{j\gamma_1}, \quad z_3-z_2 =\rho_2\ex^{j\gamma_2},
\]
where $\gamma_1$, $\gamma_2$ are the angles shown in Fig. \ref{fig:function2}.  Then
\[
H=\Im \{ (z_1-z_2)\overline{(z_3-z_2)}\} = \Im \{\rho_1\ex^{j\gamma_1}\rho_2\ex^{-j\gamma_2} \}= \rho_1\rho_2\sin(\beta_2)
\]
Thus, $0 < \beta_2 < \pi$, $\rho_1 > 0$, and $\rho_2 > 0$ iff $H >0$; and $\pi < \beta_2 < 2\pi$, $\rho_1 > 0$, and $\rho_2 > 0$ iff $H <0$.  Also, the points are collinear iff $H=0$.
\end{proof}

\begin{lemma}
\label{lem:star_convex}
If an $n$-gon is strictly convex, with its vertices $z_i$, $i=1,\ldots,n$, numbered counterclockwise around the $n$-gon, then these vertices are in a counterclockwise star formation about their centroid.
\end{lemma}
\begin{proof}
The centroid, $\tilde z$, of the $n$ vertices must lie in the interior of the strictly convex $n$-gon for if it lies on the boundary or in the exterior, we could draw a separating line through the centroid for which all vertices lie on, or to one side, contradicting the position of the centroid.  With this observation, and the fact that the $n$-gon is convex and numbered counterclockwise, the result is straightforward.
\end{proof}

With these two lemmas, and Theorem \ref{thm:star}, we can prove the main result of this section.  The result is analogous to convex curves remaining convex under \eqref{eq:euclidean_shortening}, which is shown in \cite{gage_1}.

\begin{figure}
\begin{center}
\includegraphics[width=10cm]{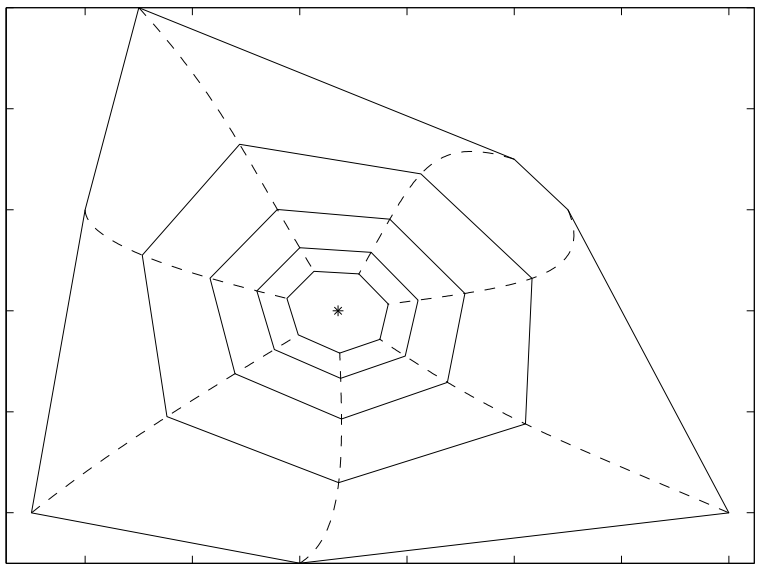}
\caption{The evolution of a convex $n$-gon.  The solid lines show the trajectories of each vertex.}
\label{fig:convex_evolution}
\end{center}
\end{figure}

\begin{theorem}
\label{thm:strictly_convex}
Consider a strictly convex $n$-gon at time $t=0$, whose vertices $z_i$, $i=1,\ldots,n$, are numbered counterclockwise.  If these vertices evolve according to \eqref{eq:evolution}, the $n$-gon will remain strictly convex for all time.
\end{theorem}
\begin{proof}
We begin by considering the function
\[
H_i(t) = \Im \{ (z_{i-1}(t)-z_i(t))\overline{(z_{i+1}(t)-z_i(t))}\} =\rho_{i-1}\rho_i\sin(\beta_i)
\]
By the definition of a strictly convex $n$-gon we have that $\rho_i(0) > 0$, and $0 < \beta_i(0) <\pi$, $\forall i$.  Hence by Lemma \ref{lem:function2}, $H_i(0) >0$, $\forall i$.  We want to show that $H_i(t) > 0$ for all $t$, which by Lemma \ref{lem:function2} shows that the $n$-gon remains strictly convex for all time.

Suppose by way of contradiction that $t_1$ is the first time that an $H_i$ becomes zero.  We can select $i=m$ such that $H_{m}(t_1) = 0$ and $H_{m+1}(t_1) >0$, for if all the $H_i$'s are zero at $t_1$, then the points are collinear, which by Lemma \ref{lem:collinear_4} is a contradiction since the points started in a convex $n$-gon formation.  Hence, we have $H_i(t) > 0$, $\forall$ $t\in[0,t_1)$ and $i=1,\ldots,n$, $H_m(t_1) = 0$, and $H_{m+1}(t_1) > 0$.

Taking the derivative of $H_m$ along the trajectories of \eqref{eq:evolution}, we have
\[
\dot H_m = \Im\{ (\dot z_{m-1} -\dot z_m)\overline{(z_{m+1}-z_m)} + (z_{m-1} -z_m)\overline{(\dot z_{m+1}-\dot z_m)} \}.
\]
Substituting in \eqref{eq:evolution} for $\dot z_{m-1},\dot z_m,\dot z_{m+1}$, and simplifying we obtain $\dot H_m = -2H_m +G_m$, where
\begin{equation}
G_m=\frac{1}{2}\Im\{(z_{m-2}-z_{m-1})\overline{(z_{m+1}-z_m)}+ (z_{m-1}-z_m)\overline{(z_{m+2}-z_{m+1})} \}.
\end{equation}
Now, if $H_m(t_1)=0$, by Lemma \ref{lem:function2}, one of the following four conditions must be satisfied. (i) $\beta_m(t_1)=\pi$ and $\rho_{m-1}(t_1),\rho_m(t_1) >0$.  (ii) $\beta_m(t_1)=0$ and $\rho_{m-1}(t_1),\rho_m(t_1) >0$. (iii) $\rho_m(t_1)= 0$. (iv) $\rho_{m-1}(t_1)= 0$.

Condition (iii), in which $\rho_m(t_1)=0$, cannot be satisfied since $H_{m+1}(t_1) >0$.  Also, since the $n$-gon is initially convex, by Lemma \ref{lem:star_convex} it is in a counterclockwise star formation.  By Theorem \ref{thm:star} the vertices remain in a star formation for all time and thus remain distinct. Therefore, condition (iv) in which $\rho_{m-1}(t_1)=0$, cannot be satisfied.

Assume condition (i) is satisfied.  Then $\beta_m(t_1)=\pi$, $\rho_{m-1}(t_1),\rho_m(t_1) >0$ and $H_m(t_1)=0$, $H_{m+1}(t_1)>0$.  Since $\beta_m(t_1)=\pi$, we have that
\[
\frac{z_{m+1}(t_1)-z_m(t_1)}{\rho_m}=-\frac{z_{m-1}(t_1)-z_m(t_1)}{\rho_{m-1}}.
\]
Combining this with the expression for $G_m$ we have
\begin{align}
G_m(t_1)&=\frac{1}{2}\Im\{-\frac{\rho_m}{\rho_{m-1}}(z_{m-2}-z_{m-1})\overline{(z_{m-1}-z_m)} -\frac{\rho_{m-1}}{\rho_{m}}(z_{m+1}-z_m)\overline{(z_{m+2}-z_{m+1})} \} \nonumber\\
\label{eq:G_m_at_t1}
&=\frac{1}{2}\left(\frac{\rho_m}{\rho_{m-1}}H_{m-1}(t_1) +\frac{\rho_{m-1}}{\rho_{m}}H_{m+1}(t_1)
\right).
\end{align}
Since $\rho_{m-1}(t_1),\rho_{m}(t_1) >0$, $H_{m+1}(t_1) >0$, and $H_{m-1}(t_1) \geq 0$, it follows that $G_m(t_1) > 0$.  By continuity of $G_m$ there exists $0\leq t_0 < t_1$ such that $G_m(t) > 0$ $\forall t\in[t_0,t_1]$.  Also, by assumption, $H_m(t) > 0$ for $t\in[0,t_1)$.  Therefore $\dot H_m(t) = -2H_m+G_m > -2H_m$, for all $t\in[t_0,t_1)$.  Integrating this and using the continuity of $H_m$, we obtain $H_m(t_1) > 0$, which is a contradiction.

Finally, assume condition (ii) is satisfied.  Then $\beta_m(t_1)=0$ and $\rho_{m-1},\rho_m >0$.  The angle $\beta_m$ is the interior angle between the edges $z_{m-1}z_m$ and $z_{m}z_{m+1}$. For all $t\in [0,t_1)$, we have $\beta_i(t)\in(0,\pi)$ and $\rho_i(t) >0$ for all $i$.  Moving $z_m$ to the origin, we can define the (positive) cone created by the edges of the $n$-gon $z_{m-1}z_m$ and $z_{m}z_{m+1}$, as $\{a(z_{m-1}-z_m)+b(z_{m+1}-z_{m}): a,b \geq 0\}$.  The $n-3$ vertices which are not involved in creating the cone must lie in the interior of this cone for all $t\in[0,t_1)$.
This is shown in Fig. \ref{fig:beta_zero}.
\begin{figure}
\begin{center}
\includegraphics[width=10cm]{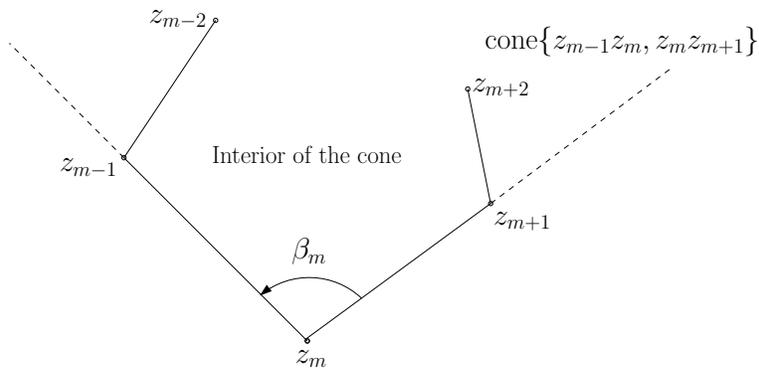}
\caption{The convex $n$-gon lying within the boundaries of the shaded cone.}
\label{fig:beta_zero}
\end{center}
\end{figure}
By continuity of the $z_i$'s, at time $t_1$ the vertices must lie either in the interior or on the boundary of this cone.  But we have $\beta_m(t_1)=0$, implying that $z_{m-1}z_m$ and $z_mz_{m+1}$ are collinear and the cone is a line.  Hence all the vertices are collinear, a contradiction by Lemma \ref{lem:collinear_4}.
\end{proof}
A straightforward consequence of the previous theorem is the following.
\begin{corollary}
\label{cor:convex}
Consider an $n$-gon which is convex at $t=0$. If the vertices evolve according to \eqref{eq:evolution}, then for any $t >0$, the $n$-gon will be strictly convex.
\end{corollary}
\begin{proof}
Consider a vertex $m$ for which $\beta_m(0)=\pi$, and thus $H_m(0)=0$.  We can choose this vertex such that $H_{m+1}(0) > 0$ since if $H_i(0)=0$, $\forall i$, then the $n$-gon is not initially convex.  From the proof of Theorem \ref{thm:strictly_convex} we have $\dot H_m(t)=-2H_m(t)+G_m(t)$.  But $H_m(0)=0$ and we have shown in \eqref{eq:G_m_at_t1} that $G_m(0) >0$.  Therefore, $\dot H_m(0) > 0$.  By continuity of $\dot H_m$ there exists a $t_0 > 0$ such that $\dot H_m(t) >0$ for $t\in[0,t_0]$. Thus, $H_m(t) >0$, for all $t\in(0,t_0]$ and by Theorem \ref{thm:strictly_convex}, $H_m(t) >0$ for all $t>t_0$.
\end{proof}
Fig. \ref{fig:convex_evolution} shows the evolution of an initially convex $n$-gon.
\section{Optimal control law for perimeter shortening}
\label{sec:optimal}

In \cite{As_fast_as_possible} it is stated that a curve evolving according to \eqref{eq:euclidean_shortening} is shrinking as fast as it can using only local information.  To see why and in what sense this is true, reparametrize the curve in terms of its Euclidean arc-length $s$, defined via the differential arc-length element $ds :=\|{\partial \mathbf{x}/\partial p}\|dp$.  With this we can write the length of a curve as
\begin{equation}
\label{eq:length_intergral}
L(t)=\int_0^{L(t)}ds = \int_0^1\left\|\frac{\partial \mathbf{x}}{\partial p}\right\| dp.
\end{equation}
In what follows we will differentiate this expression and determine the direction for the curve evolution which maximizes the rate of decrease of $L(t)$.  In order to take the time derivative of this expression, first consider taking the time derivative of $\|\partial \mathbf{x}/\partial p\|^2$:
\[
\frac{\partial }{\partial t} \left\|\frac{\partial \mathbf{x}}{\partial p}\right\|^2
= \frac{\partial }{\partial t} \left\langle \frac{\partial \mathbf{x}}{\partial p},\frac{\partial \mathbf{x}}{\partial p}  \right\rangle=2\left\langle \frac{\partial \mathbf{x}}{\partial p},
\frac{\partial }{\partial p}\frac{\partial \mathbf{x}}{\partial t}  \right\rangle,
\]
where $\langle \cdot \;, \cdot \rangle$ is the inner product (for $u,v\in\R^n$, $\langle u, v \rangle = u^Tv$).
We also have that
\[
\frac{\partial }{\partial t} \left\|\frac{\partial \mathbf{x}}{\partial p}\right\|^2=
2\left\|\frac{\partial \mathbf{x}}{\partial p}\right\|\frac{\partial }{\partial t} \left(\left\|\frac{\partial \mathbf{x}}{\partial p}\right\| \right).
\]
Therefore, combining these expressions and using the notation $\|\partial \mathbf{x}/ \partial p\|=\|\mathbf{\dot x}\|$, we have
\[
\frac{\partial }{\partial t} \left(\left\|\frac{\partial \mathbf{x}}{\partial p}\right\| \right)
= \frac{1}{\|\mathbf{\dot x}\|}\left\langle \frac{\partial \mathbf{x}}{\partial p},
\frac{\partial }{\partial p}\frac{\partial \mathbf{x}}{\partial t}  \right\rangle.
\]
Using this expression in \eqref{eq:length_intergral} we obtain
\[
\frac{d L}{dt}= \int_0^1 \frac{1}{\|\mathbf{\dot x}\|}\left\langle \frac{\partial \mathbf{x}}{\partial p},
\frac{\partial }{\partial p}\frac{\partial \mathbf{x}}{\partial t}  \right\rangle dp.
\]
Now,
\[
\frac{1}{\|\mathbf{\dot x}\|}\frac{\partial \mathbf{x}}{\partial p}=\frac{1}{\|\mathbf{\dot x}\|}
\frac{\partial s}{\partial p} \frac{\partial \mathbf{x}}{\partial s}=\frac{\partial \mathbf{x}}{\partial s},
\]
since $ds =\|\mathbf{\dot x}\|dp$.  This gives us
\[
\frac{d L}{dt}= \int_0^1 \left\langle \frac{\partial \mathbf{x}}{\partial s},
\frac{\partial }{\partial p}\frac{\partial \mathbf{x}}{\partial t}  \right\rangle dp.
\]
Integrating by parts we obtain
\[
\frac{d L}{dt}= \left. \left\langle \frac{\partial \mathbf{x}}{\partial s}, \frac{\partial \mathbf{x}}{\partial t}\right\rangle \right|_{0}^1 -
\int_0^1 \frac{\partial }{\partial p} \left(\frac{\partial \mathbf{x}}{\partial s}\right)^T \frac{\partial \mathbf{x}}{\partial t}  dp.
\]
The first term on the right-hand side is zero since the curve is smooth and $\mathbf{x}(0,t)=\mathbf{x}(1,t)$, and the second term can be rewritten to obtain
\[
\frac{d L}{dt}= -
\int_0^L \left(\frac{\partial^2 \mathbf{x}}{\partial s^2}\right)^T \frac{\partial \mathbf{x}}{\partial t} ds.
\]
Finally, since $\partial \mathbf{x}/\partial s= \mathbf{T}$ and $\partial \mathbf{T}/\partial s = k\mathbf{N}$, we have
\begin{equation}
\label{eq:length_decrease}
\frac{d L}{dt}=-\int_0^L \left\langle k \mathbf{N}, \frac{\partial \mathbf{x}}{\partial t} \right\rangle ds.
\end{equation}
Therefore, the direction of $\partial \mathbf{x}/\partial t$ in which $L(t)$ is decreasing most rapidly is $\partial \mathbf{x}/\partial t = k\mathbf{N}$, which is the Euclidean curve shortening rule \eqref{eq:euclidean_shortening}. Note that this flow is optimal only in the sense that, given a fixed
magnitude of the velocity of the curve at each point, this velocity always points in the direction which maximizes the rate of decrease of $L(t)$.

\begin{figure}
\begin{center}
 \subfigure[A polygon evolving in the optimal direction.]
  {\label{fig:optimal_direction}\includegraphics[width=7.5cm]{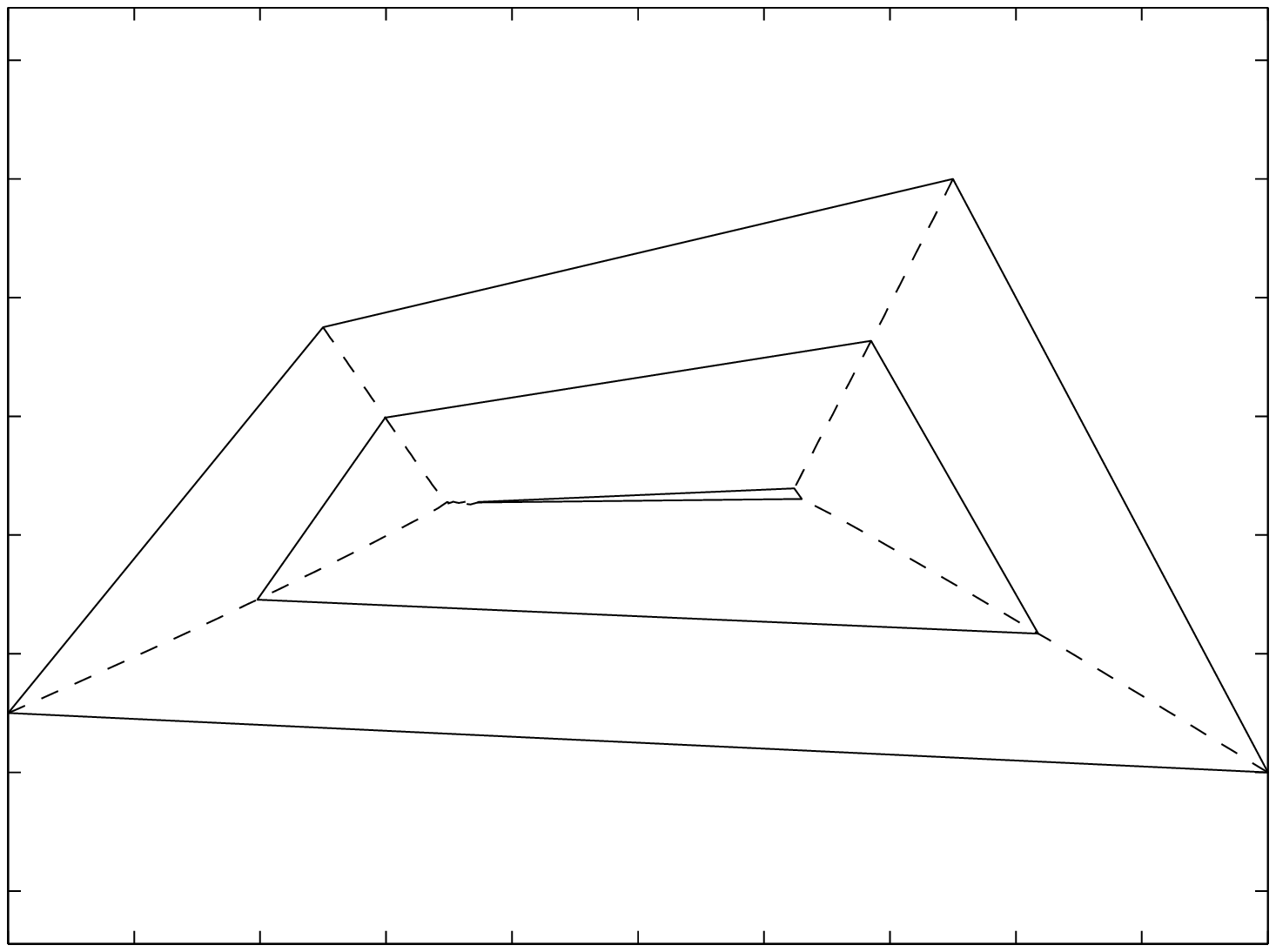}}
\hfill
\subfigure[The same polygon evolving according to linear polygon shortening.]
{\label{fig:optimal_direction_compare} \includegraphics[width=7.5cm]{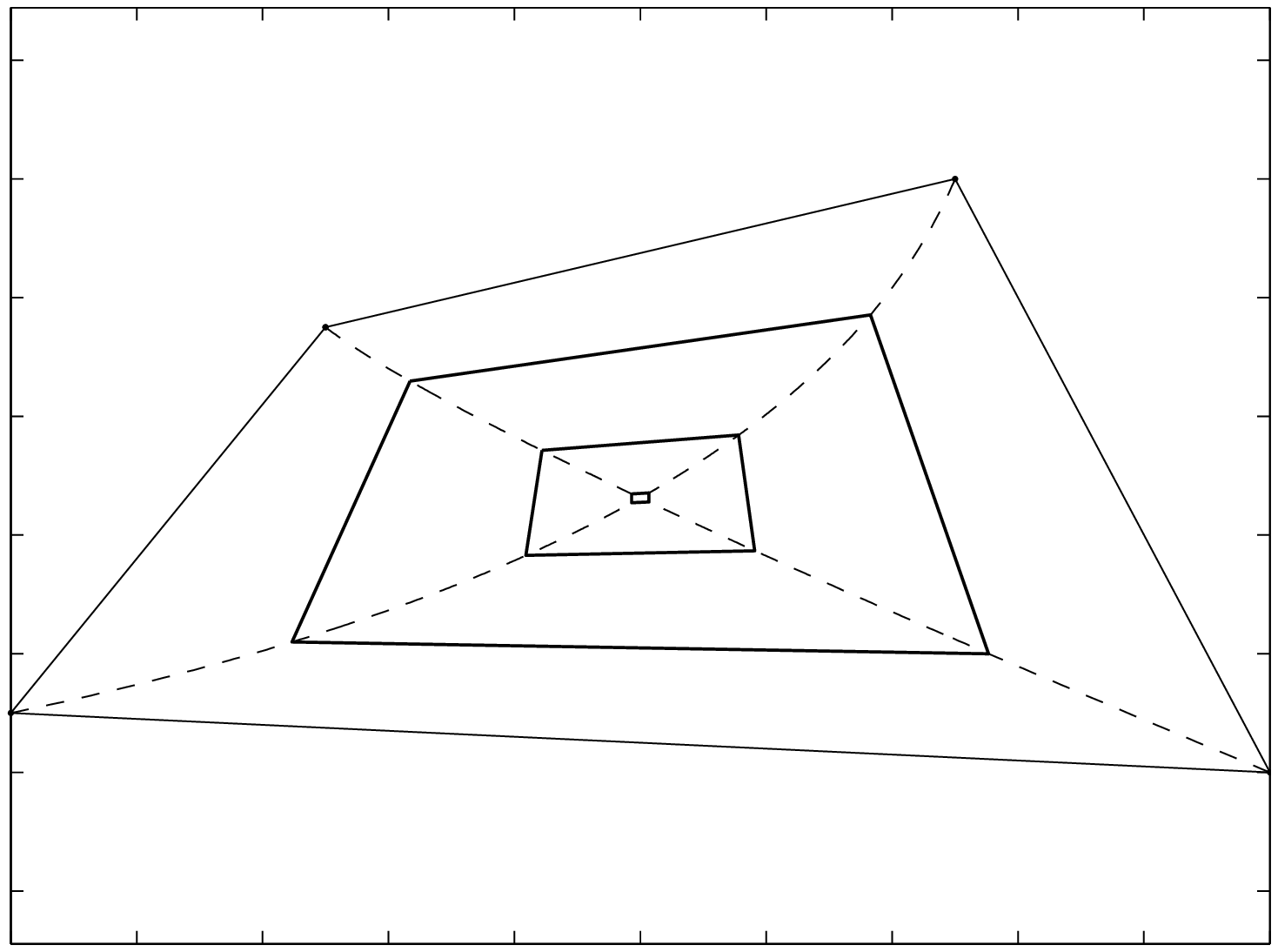}}
\caption{Evolving in the optimal direction.}
\label{fig:optimal}
\end{center}
\end{figure}

We now give an analogous result for the discrete polygon case.
Given an $n$-gon we can write its perimeter as
\begin{equation}
\label{eq:perimeter}
P(t)=\sum_{i=1}^n|z_{i+1}-z_i|.
\end{equation}
In order to take the time derivative of $P(t)$ consider taking the derivative of $|z_{i+1}-z_i|^2 = \langle z_{i+1}-z_i,z_{i+1}-z_i \rangle$ (for $u,v\in\mathbb{C}^n, \langle u,v \rangle =u^*v$, where $^*$ denotes complex conjugate transpose).
This yields
\begin{align*}
\frac{d}{dt}|z_{i+1}-z_i|^2 &= \frac{d}{dt}\langle z_{i+1}-z_i,z_{i+1}-z_i \rangle \\
&=2\Re\left\{\langle z_{i+1}-z_i,\dot z_{i+1}-\dot z_i \rangle\right\}.
\end{align*}
But also,
\[
\frac{d}{dt}|z_{i+1}-z_i|^2=2 |z_{i+1}-z_i| \frac{d}{dt}|z_{i+1}-z_i|.
\]
Letting $\dot z_i=u_i$ for $i=1,\ldots,n$ and rearranging we have
\[
\frac{d}{dt}|z_{i+1}-z_i|=\Re\left\{\left\langle
\frac{z_{i+1}-z_i}{|z_{i+1}-z_i|},u_{i+1}-u_i\right\rangle \right\}.
\]
Therefore
\[
\dot P(t)=\sum_{i=1}^n\Re\left\{\left\langle
\frac{z_{i+1}-z_i}{|z_{i+1}-z_i|},u_{i+1}-u_i\right\rangle \right\}.
\]
Since all indices are evaluated modulo $n$ this can be rewritten as
\begin{equation}
\label{eq:P(t)_inner_product}
\dot P(t)= -\sum_{i=1}^n\Re\left\{\left\langle \frac{z_{i-1}-z_{i}}{|z_{i-1}-z_{i}|}+
\frac{z_{i+1}-z_i}{|z_{i+1}-z_i|},u_{i} \right\rangle \right\}.
\end{equation}
To maximize the rate of decrease of $P(t)$, $u_i$ should point in the direction of $(z_{i-1}-z_{i})/|z_{i-1}-z_{i}| + (z_{i+1}-z_i)/|z_{i+1}-z_i|$.  This direction bisects the internal angle $\beta_i$ of the $n$-gon.  In general, neither the linear scheme \eqref{eq:evolution} nor the shortening by Menger-Melnikov curvature points in this direction.  However, this direction does not ensure that the polygon becomes circular (nor elliptical); in simulation, adjacent vertices may capture each other and the polygon may collapse to a line.An example is shown in Fig. \ref{fig:optimal}.  Notice that these undesirable features do not appear in the linear scheme.

Using \eqref{eq:P(t)_inner_product} and \eqref{eq:evolution} we can determine $\dot P(t)$.  For $\dot P(t)$ to be defined we require that adjacent vertices be distinct.  This is ensured, for example, if the vertices start in a star formation about their centroid.  The following result is analogous to the result in \cite{gage_1} that under \eqref{eq:euclidean_shortening} the length of the curve monotonically decreases.
\begin{theorem}
Consider an $n$-gon whose distinct vertices evolve according to \eqref{eq:evolution}.  If adjacent vertices remain distinct, the perimeter $P(t)$ of the $n$-gon monotonically decreases to zero.
\end{theorem}
{\em Proof:} \
Substituting \eqref{eq:evolution} into \eqref{eq:P(t)_inner_product} and expanding we obtain
\begin{eqnarray*}
\dot P(t) & = & \frac{1}{2}\sum_{i=1}^n\Re\left\{-|z_{i}-z_{i-1}|-|z_{i+1}-z_i|+
\left\langle \frac{z_{i}-z_{i-1}}{|z_{i}-z_{i-1}|},z_{i+1}-z_i\right\rangle \right.\\
& & +\left.\left\langle \frac{z_{i+1}-z_i}{|z_{i+1}-z_i|},z_{i}-z_{i-1} \right\rangle
\right\}.
\end{eqnarray*}
Each term in this summation has the form $\Re\{-|u|-|v|+\langle u/|u|,v\rangle + \langle v/|v|,u\rangle \}$.  From the Cauchy-Schwarz inequality we have $\Re\{\langle u/|u|,v\rangle\} \leq |v|$, $\Re\{\langle v/|v|,u\rangle\} \leq |u|$,  and thus $\Re\{-|u|-|v|+\langle u/|u|,v\rangle + \langle v/|v|,u\rangle \}\leq 0$.  Therefore, $\dot P(t) \leq 0$.  Equality is achieved if and only if $u/|u|=v/|v|$ for each term in the summation; that is, if and only if
\begin{equation}
\label{eq:summation_maximum}
\frac{z_{i}-z_{i-1}}{|z_{i}-z_{i-1}|}=\frac{z_{i+1}-z_i}{|z_{i+1}-z_i|}, \quad \forall i.
\end{equation}
However, assume by way of contradiction that \eqref{eq:summation_maximum} is satisfied.  Rotate the coordinate system such that $z_1$ and $z_2$ lie on the real axis and $z_2 - z_1 >0$.  Setting $i=2$ in \eqref{eq:summation_maximum} we have $z_3-z_2>0$, setting $i=3$ we have $z_4-z_3>0$, and so on.  Hence $z_{i+1}-z_i >0$, $\forall i=1,\ldots,n-1$, which implies that $z_n >z_1$.  But setting $i=n$ in \eqref{eq:summation_maximum} we have $z_1-z_n >0$, a contradiction.  Therefore \eqref{eq:summation_maximum} cannot be satisfied, $\dot P(t)<0$, and since the vertices converge to their stationary centroid, $P(t)$ monotonically decreases to zero. \hfill $\Box$

\section{Limitations of the linear scheme}
\label{sec:limitations}
\begin{figure}
\begin{center}
 \subfigure[The evolution of a simple polygon.  The dashed lines show the trajectories of the vertices.]
  {\label{fig:boomerang}\includegraphics[width=7.5cm]{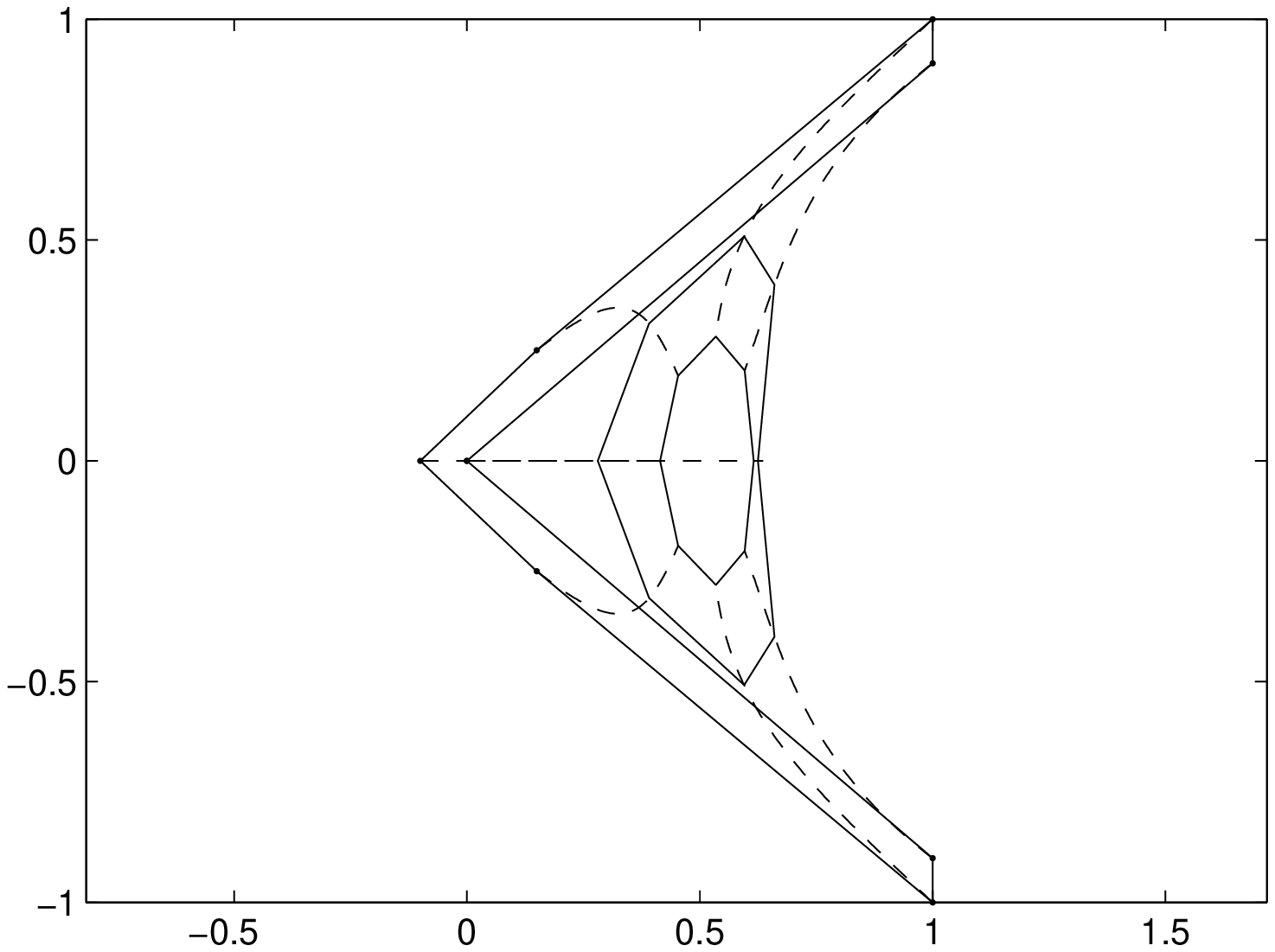}}
\hfill
\subfigure[A plot of the area as a function of time.  Notice that the area is initially increasing.]
{\label{fig:area_vs_time}\includegraphics[width=7.5cm]{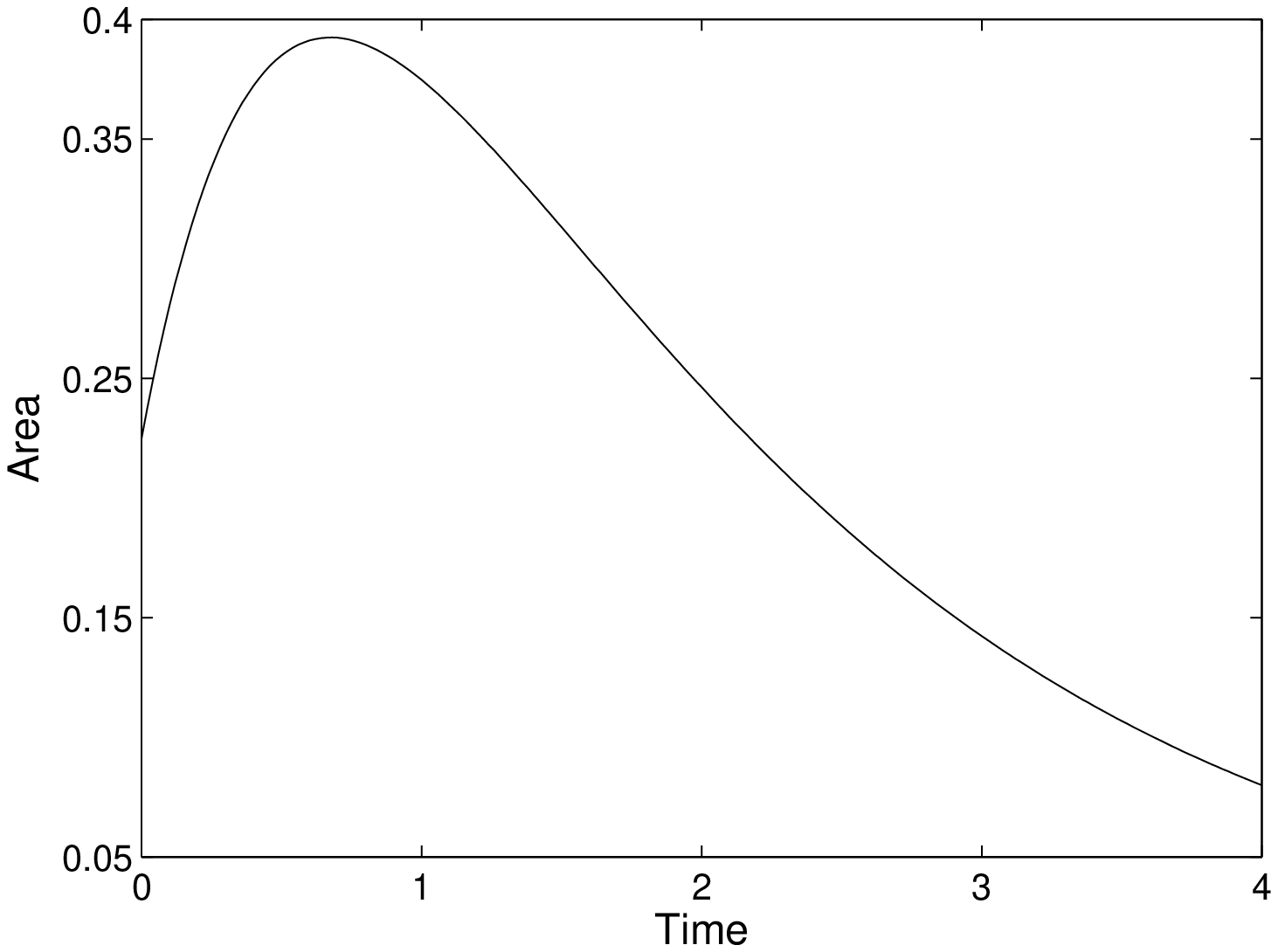}}
\caption{An embedded polygon for which the area initially increases.}
\label{fig:increasing area}
\end{center}
\end{figure}

There are two ways in which the linear scheme does not mimic Euclidean curve shortening.  First of all, if an embedded curve is evolved via Euclidean curve shortening, its area is monotonically decreasing.  However, for the linear scheme, in general, the area of a simple polygon is not monotonically decreasing.
This is shown in Fig. \ref{fig:increasing area}.
An interesting observation is that if a convex polygon evolves according to \eqref{eq:evolution}, its area is monotonically decreasing.  To see this, consider a convex polygon at time $t=0$ with vertices $z_i$, $i=1,\ldots,n$, evolving according to \eqref{eq:evolution}.  For each $i$, $\dot z_i(0)$ is either zero, or points into the interior of the polygon, with $\dot z_i(0) \neq 0$ for some $i$.  Therefore, the area is initially decreasing.  By Corollary \ref{cor:convex} the polygon is strictly convex for all $t >0$, and thus $\dot z_i(t)$ points into the interior of the polygon for all $i$ and for all $t>0$.  Therefore, the area decreases for all time.

\begin{figure}
\begin{center}
 \subfigure[A simple polygon. The vertices are marked by $*$'s.]
  {\label{fig:embedded_initial}\includegraphics[width=7.5cm]{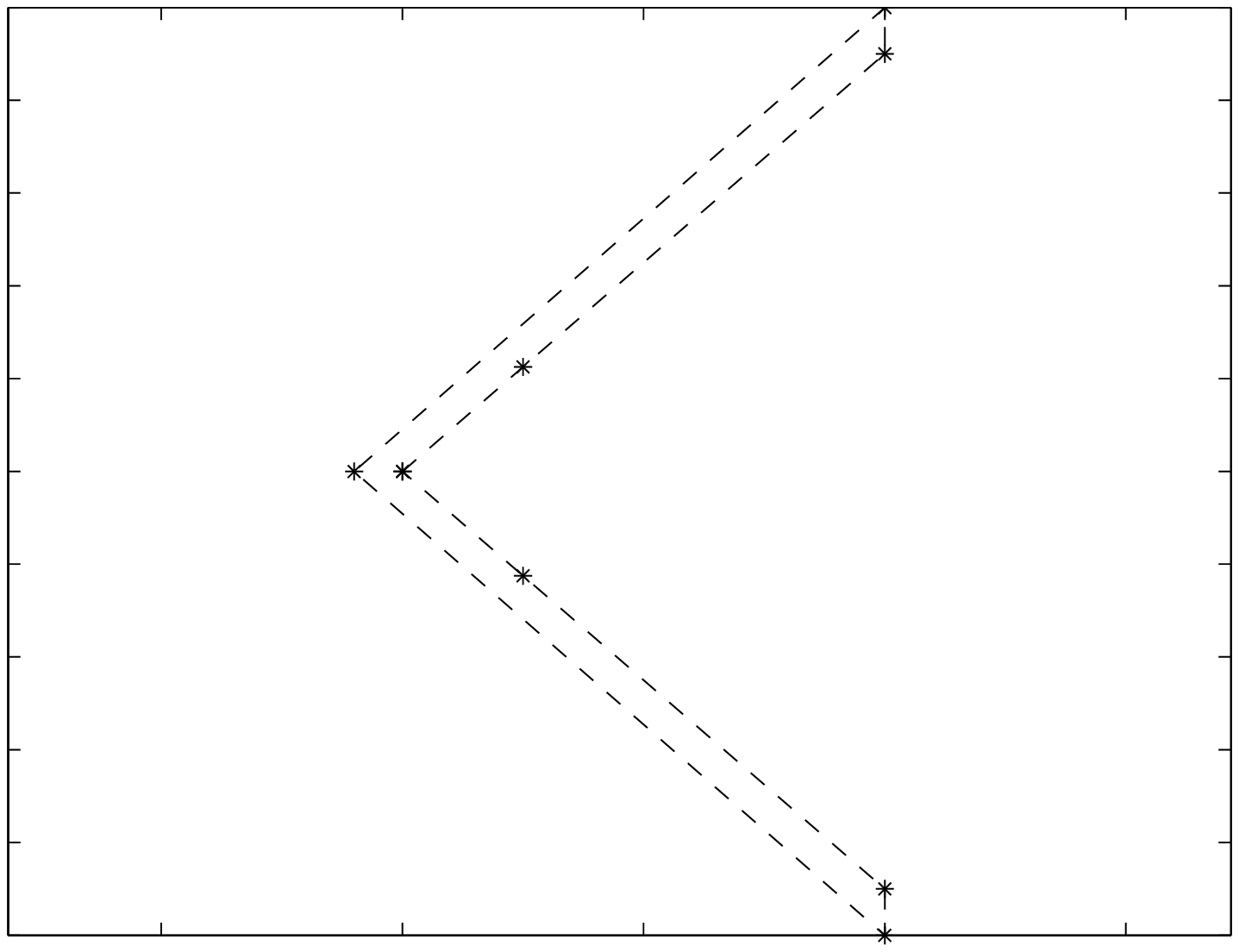}}
\hfill
\subfigure[The initial polygon evolves to the self-intersecting polygon shown by the thick solid line.]
{\label{fig:embedded_final}\includegraphics[width=7.5cm]{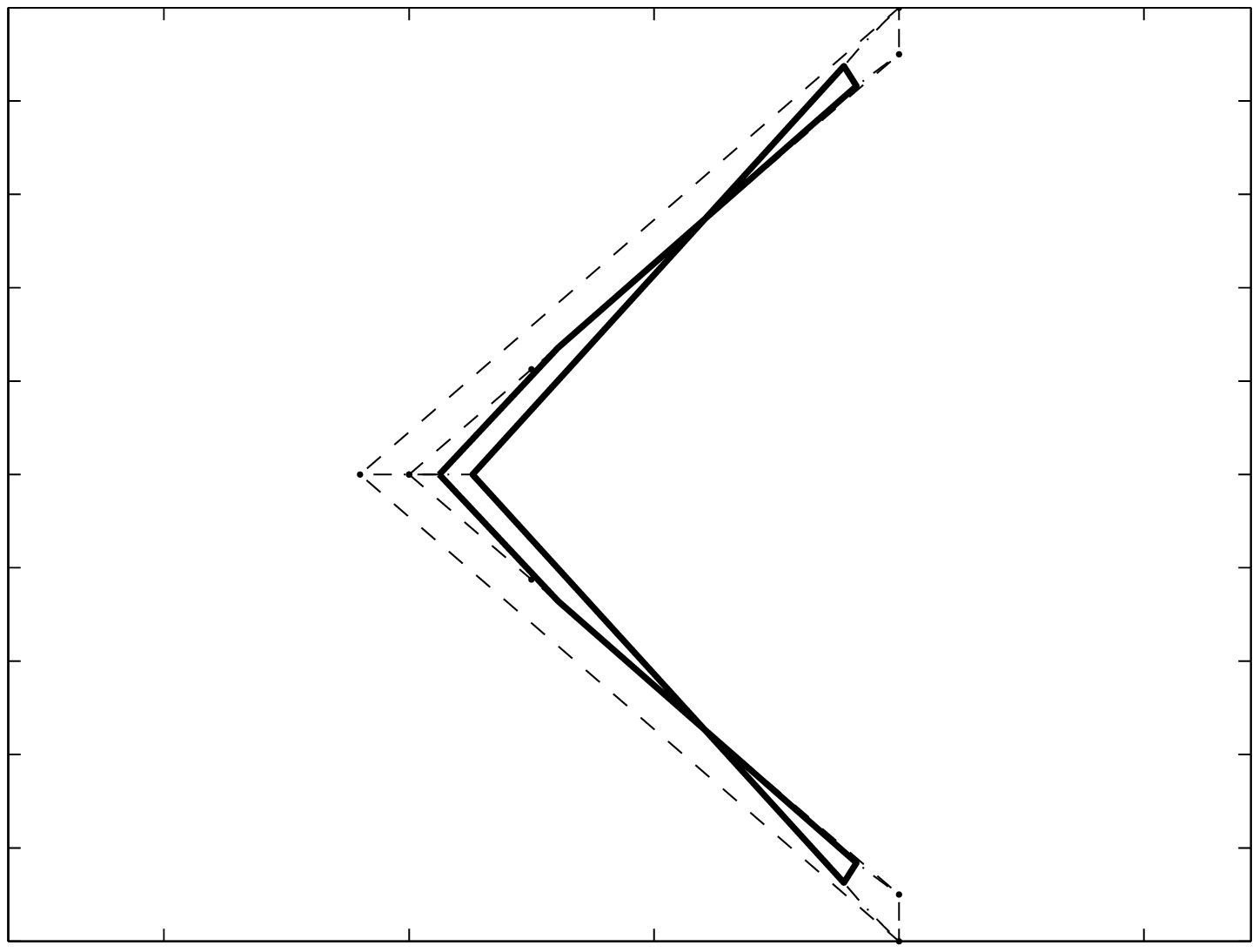}}
\caption{A simple polygon becomes self-intersecting.}
\label{fig:embedded}
\end{center}
\end{figure}

The second way in which the linear scheme does not mimic Euclidean curve shortening is in its effect on simple $n$-gons. If an embedded curve evolves according to the Euclidean curve shortening flow, it remains embedded.  In contrast, a simple $n$-gon can become self-intersecting under the linear scheme, as is shown in Fig. \ref{fig:embedded}.
However, this is to be expected since the vertices in Fig. \ref{fig:embedded_initial} are not equally spaced around the polygon.  The regions of the polygon with smaller spacing between adjacent vertices will move more slowly than the regions where the spacing is large.  This is why, in Fig. \ref{fig:embedded_final}, the outer edge of the boomerang has intersected the inner edge.

\section{Conclusion}

In summary, under the simple distributed linear control law \eqref{eq:evolution}, the robots rendezvous and also become more organized, in the sense that the polygon becomes elliptical. Furthermore, star formations remain so, convex polygons remain so, and the perimeter of the polygon decreases monotonically. These results are intended as a possible starting point for more useful behavior. As an example scenario, consider a number of mobile robots initially placed at random, and which should self-organize into a regular polygon (circle) for the purpose of forming a large-aperture antenna. Distributed control laws for this goal would have to be nonlinear. Research on this front is on-going.

Another topic for future research is to look at polygon shortening flows for wheeled robots which are subject to nonholonomic motion constraints.

Finally, drawing upon the results on curve shortening flows, there has been a similar development of curve expanding flows---If a smooth, closed, and embedded curve is deformed along its \emph{outer} normal vector field at a rate proportional to the \emph{inverse of its curvature}, it expands to infinity, and the limiting shape is circular \cite{curve_expansion}.  Thus, a scheme for \emph{deployment} of a fleet of mobile robots could be achieved by creating the analogous polygon expanding flow.
\bibliography{arxiv}

\end{document}